\documentclass[10pt,twocolumn,letterpaper]{article}
\usepackage{pgfplots}
\pgfplotsset{compat=1.16}
\usepackage{tikz}

\usepackage{iccv}
\usepackage{times}
\usepackage{epsfig}
\usepackage{graphicx}
\usepackage{amsmath}
\usepackage{amssymb}

\usepackage{graphicx}
\usepackage{amsmath}
\usepackage{amssymb}
\usepackage{booktabs}

\usepackage{subfigure}
\usepackage[export]{adjustbox}

\usepackage[pagebackref=true,breaklinks=true,letterpaper=true,colorlinks,bookmarks=false]{hyperref}
\usepackage{iccv}
\usepackage{times}
\usepackage{epsfig}
\usepackage{graphicx}
\usepackage{amsmath}
\usepackage{amssymb}
\usepackage[accsupp]{axessibility}
\usepackage{authblk}

\usepackage[breaklinks=true,bookmarks=false]{hyperref}

\iccvfinalcopy 

\def\iccvPaperID{****} 

\ificcvfinal\fi

\title{Most Important Person-guided Dual-branch Cross-Patch Attention for Group Affect Recognition}

\author[1]{Hongxia Xie}
\author[2]{Ming-Xian Lee}
\author[2]{Tzu-Jui Chen}
\author[1]{Hung-Jen Chen}
\author[1]{Hou-I Liu}
\author[1]{Hong-Han Shuai}
\author[3]{Wen-Huang Cheng}

\affil[1]{National Yang Ming Chiao Tung University, \texttt{hongxiaxie.ee08@nycu.edu.tw}}
\affil[2]{University of Illinois Urbana-Champaign}
\affil[3]{National Taiwan University, \texttt{wenhuang@csie.ntu.edu.tw}}

\begin{document}

\maketitle
\ificcvfinal\thispagestyle{empty}\fi


\begin{abstract}
Group affect refers to the subjective emotion that is evoked by an external stimulus in a group, which is an important factor that shapes group behavior and outcomes. Recognizing group affect involves identifying important individuals and salient objects among a crowd that can evoke emotions. 
However, most existing methods lack attention to affective meaning in group dynamics and fail to account for the contextual relevance of faces and objects in group-level images.
In this work, we propose a solution by incorporating the psychological concept of the Most Important Person (MIP), which represents the most noteworthy face in a crowd and has affective semantic meaning.
We present the Dual-branch Cross-Patch Attention Transformer (DCAT) which uses global image and MIP together as inputs.
Specifically, we first learn the informative facial regions produced by the MIP and the global context separately. Then, the Cross-Patch Attention module is proposed to fuse the features of MIP and global context together to complement each other. 
Our proposed method outperforms state-of-the-art methods on GAF 3.0, GroupEmoW, and HECO datasets. Moreover, we demonstrate the potential for broader applications by showing that our proposed model can be transferred to another group affect task, group cohesion, and achieve comparable results.

\end{abstract}

\section{Introduction}
Humans are active and social creatures, using multimodal interactions to convey their intentions, attitudes, and feelings.
Such physical and emotional interactions between individuals within a group can generate group-level affect or group-level emotion~\cite{groupaffect}~\footnote{
In this work, groups range in size from two or three individuals up to hundreds.}.
\begin{table}[t]
\small
\begin{center}
\begin{tabular}{llllll}
\toprule
  \textbf{Global +}  & All & 3Random  & 3MIPs & 1Random & MIP\\
\midrule
\textbf{Acc (\%)} & 85.01 & 85.47 & 87.46  & 82.30 &\textbf{88.86} \\
\bottomrule
\end{tabular}
\end{center}
\caption{Different face selection strategies. The ``3Random" refers to selecting 3 faces randomly from the detected faces in the image, while the ``3MIPs" strategy selects the top three detected faces based on their importance ranking. The dataset evaluated here is the GroupEmoW~\cite{groupemow_sota_wacv21}.}
\label{tab:MIP}
\end{table}

\begin{figure}[htb]
\begin{center}
\includegraphics[scale=0.4]{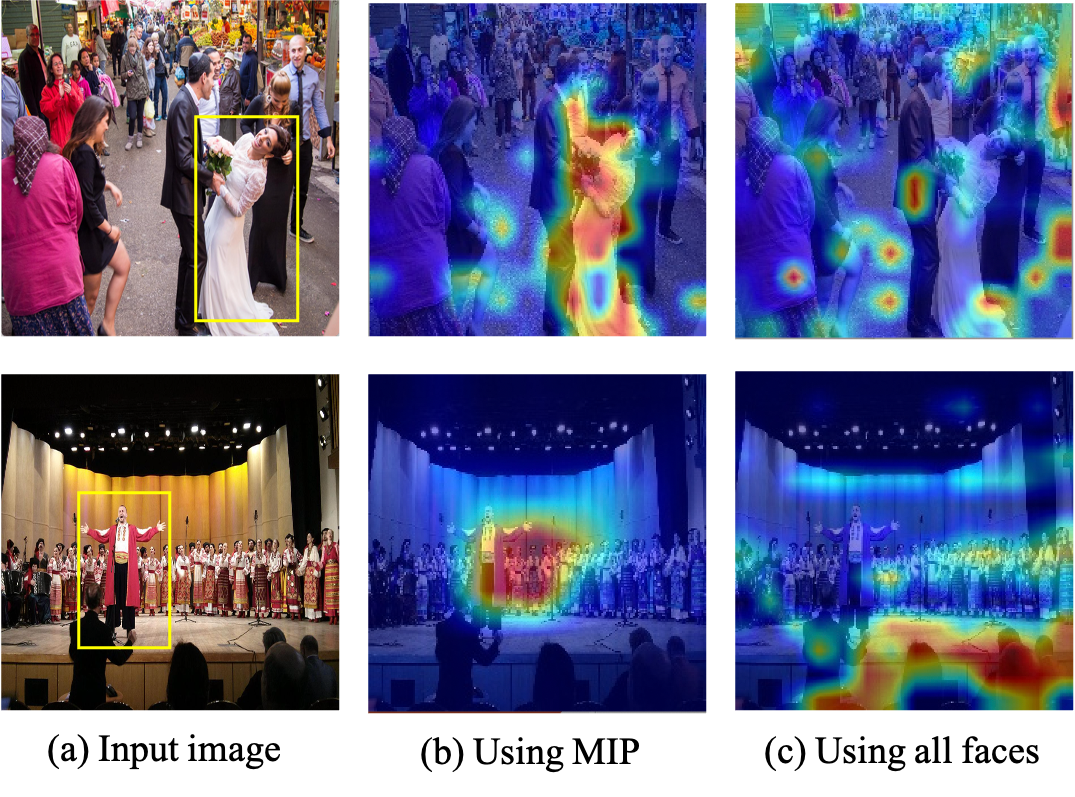}
  \caption{Feature map comparison between using MIP and all faces. The yellow box in (a) represents the MIP region.
}
  \label{fig:saliency_mip}
  \end{center}
\end{figure}

By providing the group-level information, group affect prediction has various real-world applications, \textit{e.g.}, work team outcome prediction~\cite{kelly2001mood}, social relationship recognition~\cite{polo2016group}, human-machine interaction systems~\cite{knapp2011physiological}. As a result, the research topics surrounding group affect are diverse, including categorical analysis, \textit{e.g.}, group valence (positive, negative, neutral) prediction~\cite{ GroupEmoW_wacv20,groupemow_sota_wacv21}, continuous intensity estimation, \textit{e.g.}, group cohesion prediction~\cite{cohesion_tac}. 

Group affect is influenced by a combination of a group’s \textit{affective context} (\textit{e.g.}, salient objects in the funeral, party) and \textit{affective composition} (\textit{i.e.}, the combination of a group members’ state and trait affect)~\cite{groupaffect}.
Previous methods for group affect recognition mainly adopted a bottom-up approach~\cite{tmm, ECCV2022_HECO, GroupEmoW_wacv20, groupemow_sota_wacv21}, where features were extracted from individuals separately and then fused to output group-level emotion. This was done by utilizing multi-cues from separate pre-trained detectors, such as facial expressions of individuals, object proposals, and scene features.
While these cues can provide valuable information, they may not fully capture the complex interplay of emotions within a group. 
As indicated in \cite{saliency_importa}, current group-level recognition architectures tend to focus on detecting salient regions, but they may lack an understanding of the affective meaning in the image. 
Therefore, there is a need for a more holistic approach that considers the most important region in the crowd and how they interact with the scene. 

Hence, in this work, we introduce a crucial psychological concept, namely, \textbf{Most Important Person (MIP)}~\cite{groupaffect}, which goes beyond saliency and considers the importance of individuals for group affect recognition.
 The MIP of an image is often the group leaders who can influence the emotion of a group~\cite{groupleader_1, groupleader_2,qi2021zero}. MIP has been explored in many related works in psychology and affective computing, such as the \textit{Affective Transfer Process} phenomenon in psychology~\cite{groupaffect}, which indicates that the MIP of an image can influence the group's emotion. 
 We conducted a preliminary experiment based on the CrossVit~\cite{crossvit} model to demonstrate the effectiveness of the MIP in group affect. Our architecture uses both the global image and faces as two inputs. As depicted in Table~\ref{tab:MIP}, the results demonstrate that using all faces is considerably less effective than using the MIP, and even less effective than randomly selecting three facial images. Additionally, as observed in the feature maps illustrated in Figure~\ref{fig:saliency_mip}, the model that employs all faces fails to predict accurately due to its focus on irrelevant background regions or faces. In contrast, the network that utilizes MIP can efficiently focus on relevant facial regions, thereby enhancing prediction accuracy. 
This is mainly due to the fact that in crowded scenes with many blurry faces, the model trained on all faces exhibits dispersed attention, while the model trained on the MIP region concentrates specifically on the relevant area. 

Based on the above observations, we present a dual-pathway vision transformer model, the Dual-branch Cross-Patch Attention Transformer (DCAT), including a global branch and a MIP branch.
With the dual-pathway, spatial-wise discriminative clues can be discovered both globally and locally in conjunction with self-attention learning.
For the dual-pathway interaction, since there are many interferences in the group, we propose the Cross-Patch Attention (CPA) module. CPA first utilizes the Token Ranking Module to select the important tokens in each path, and then the cross-attention is calculated between the selected query vectors and the entire key-value vectors from the other path.
In this way, global and MIP contexts are able to complement and compensate for each other.

This paper makes the following main contributions:
\begin{itemize}
\item In this work, we validate that MIP plays a crucial role in group affect recognition. To the best of our knowledge, this is the first work introducing MIP into the group affect task.
\item
The MIP and global affective context information are integrated into the proposed dual-pathway vision transformer architecture for recognizing group emotions in a context-aware manner. We also propose a novel CPA module that focuses on the interactions among diverse scene contexts and facial informative regions.
\item Experimental results show that the proposed DCAT outperforms both state-of-the-art group affect works and vision transformer models in terms of accuracy and model parameters. This is also the first work to explore group affect recognition that goes beyond the CNN-based approach. Moreover, our proposed model can be utilized in other group-level affect tasks, \textit{i.e.}, group cohesion analysis. 
\end{itemize}


\section{Related works}
\subsection{Group affect recognition methods}
Current group affect recognition methods primarily analyze the individual members of a group and then assess their contribution to the overall mood of the group by taking into account a variety of cues, such as faces, salient objects, and scenes.
Fujii \textit{et al.}~\cite{tmm} proposed a two-stage classification method, in which the first stage classifies facial expressions, and the second stage adds scene features into consideration to fuse faces and scenes through spectral clustering for further group affect recognition.
Khan \textit{et al.}~\cite{groupemow_sota_wacv21} considered multi-cues, which include all faces, object proposals, and scene information. To achieve such an architecture, additional face detection and object proposal detection detectors are needed. 
Similarly, Guo \textit{et al.}~\cite{GroupEmoW_wacv20} also investigated multi-cues using graph neural networks, where the topology of the graph varies as the number of entities varies.


However, in large groups, it is not feasible to consider the expressions of all faces and objects individually before aggregating, resulting in lower efficiency and accuracy.
As indicated in~\cite{tmm}, considering the performance of the main subject estimation is needed, while our work fills the gap.


\subsection{Most important person (MIP) detection}

\subsubsection{MIP detection}
A MIP detection model understands a high-level pattern in a social event image to identify the most important person.
Ramanathan \textit{et al.}~\cite{MIP_ramanathan2016detecting} used an RNN to learn time-varying attention weights that can be used for event detection and classification. Without any explicit annotations, the person with the highest attendance can be identified as the MIP.
In order to overcome the inherent variability of human poses, PersonRank~\cite{MIP_li2018personrank} produced a multiple Hyper-Interaction Graph, which treats each individual person as a node. The most active node can be identified using four types of edge message functions. 
POINT~\cite{POINT} proposed two types of interaction modules, \textit{i.e.}, the person-person interaction module and the event-person interaction module. It then aggregated relation features and person features to form the important feature.
To reduce the annotation effort, Hong \textit{et al.}~\cite{MIP_hong2020learning} formulated a POINT-based iterative learning method for semi-supervised important people detection.
Thus, we incorporate people awareness into global-level affect analysis in this work.
\subsection{Dual attention learning models}

Visual attention with dual-pathway has been widely used to find important contextual regions. 
Based on the outstanding performance of vision transformers (ViTs) in various visual tasks~\cite{vit}, there are many self-attention-based dual-pathway networks in recent years.
Zhu \textit{et al.}~\cite{zhu2022dual} proposed a global-local cross-attention to reinforce the spatial-wise discriminative clues for fine-grained recognition. A pair-wise cross-attention was designed to establish the interactions between image pairs.
Dual-ViT~\cite{yao2022dual} incorporated a compressed semantic pathway that served as prior information in learning finer pixel-level details.
The semantic pathway and pixel pathway are integrated together and trained to spread enhanced self-attention information in parallel.

The most similar architecture to our work is the \textbf{Cross-ViT}~\cite{crossvit}, which was proposed to learn multi-scale features effectively.
However, we differ from it in two main ways. Firstly, our network divides at different scales using global images and MIP images, whereas CrossViT divides at different scales using the same input image. Another difference is the cross-attention mechanism. The CrossViT exchanges class tokens, while we designed a Cross-Patch Attention module based on attentive queries from both paths.


\section{Proposed Dual-branch Cross-Patch Attention Transformer (DCAT)}

\begin{figure*}
  \centering
    \includegraphics[scale=0.065]{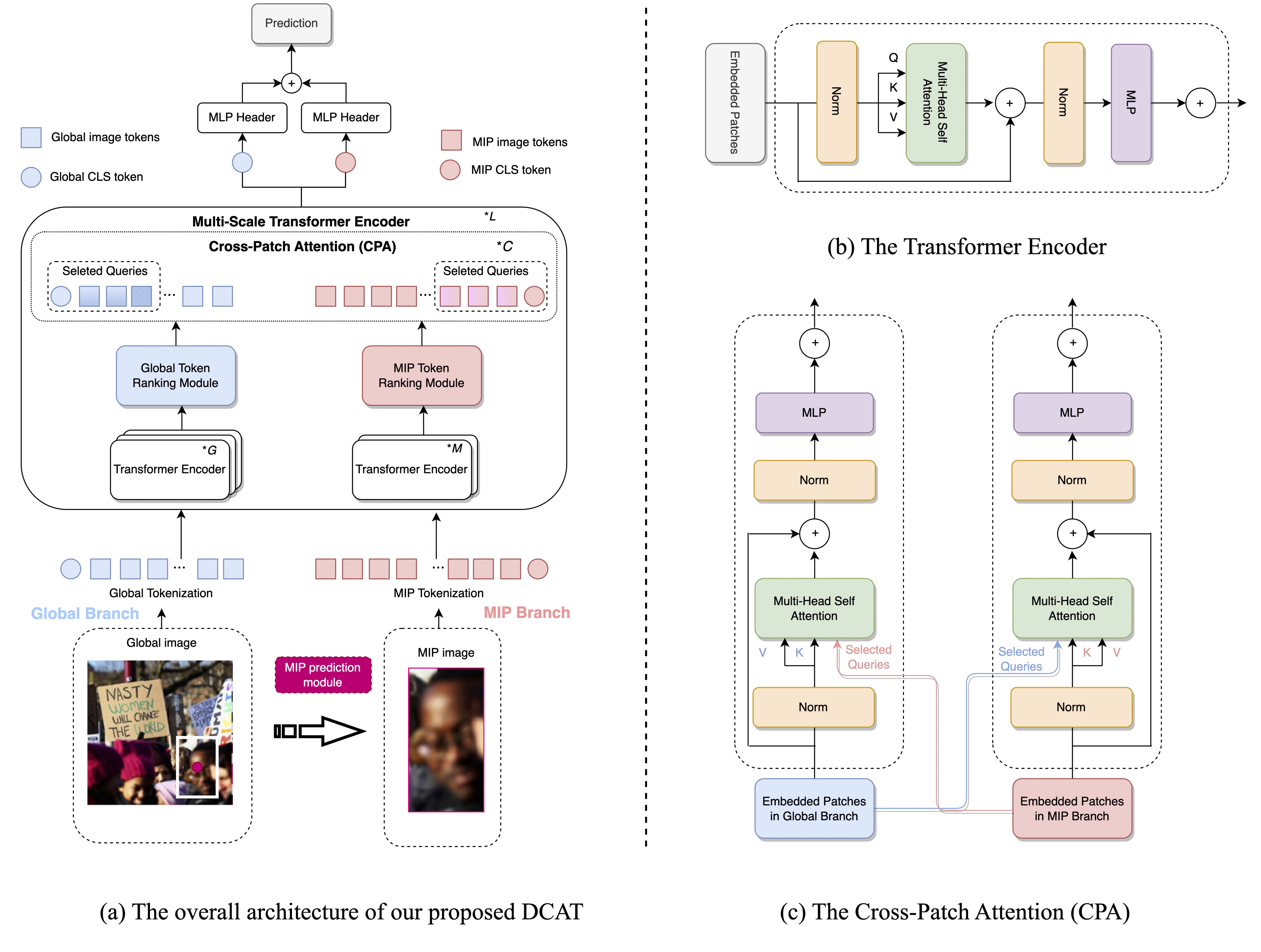}
  \caption{The overall architecture of our proposed Dual-branch Cross-Patch Attention Transformer (DCAT).}
  \label{fig:overall architecture}
\end{figure*}
\subsection{Overview}
The architecture of our proposed Dual-branch Cross-Patch Attention Transformer (DCAT) for group affect recognition is shown in Figure~\ref{fig:overall architecture}.
The model contains two inputs, \textit{i.e.}, the global image and the corresponding MIP image.
The two images are tokenized into patches first and sent into a stack of Multi-Scale Transformer Encoder, which consists of dual-path Transformer Encoder, Token Ranking Module, and Cross-Patch Attention (CPA).
Specifically, in a dual-path Transformer Encoder, the global image is sent to the coarse-level path to capture long-range information and further refine the encoded tokens for fine-grained details. On the other hand, the MIP image is fed to the fine-level path to obtain high-level semantic tokens. 
Next, since the global and MIP features are conceptual interdependent, we incorporate dual-path interaction in our model.
In particular, we sort tokens according to their importance in each path based on the Token Ranking Module, then calculate the CPA using the key and value vectors from the other path.
Finally, the class tokens from two branches are combined and sent into a linear layer for prediction.




\subsection{Dual-pathway learning}
\label{sec:dual-pathway learning}


\begin{figure}[htbp]
  \centering
  \begin{minipage}[t]{0.48\linewidth}
    \centering
    \includegraphics[width=\linewidth]{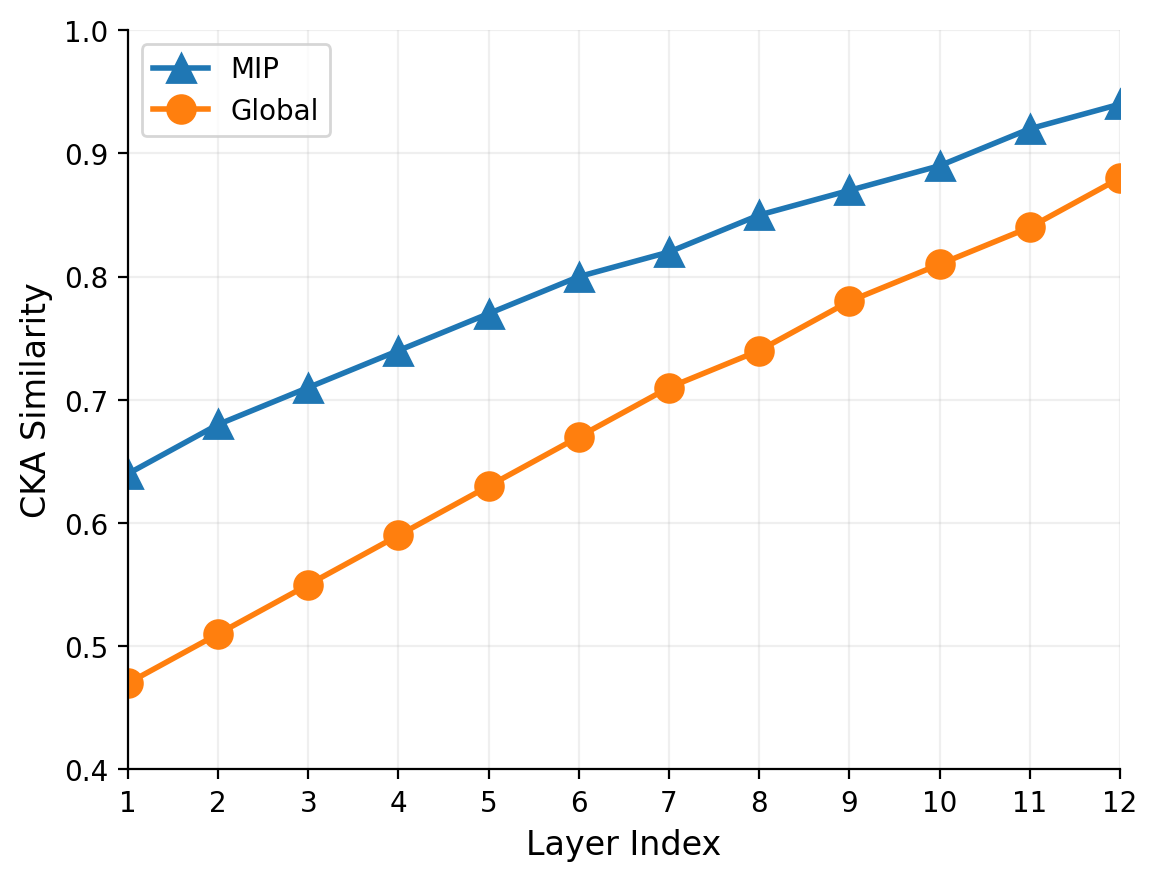}
    \caption{CKA similarity in each layer.}
    \label{fig:CKA}
  \end{minipage}
  \hfill
  \begin{minipage}[t]{0.48\linewidth}
    \centering
    \includegraphics[width=\linewidth]{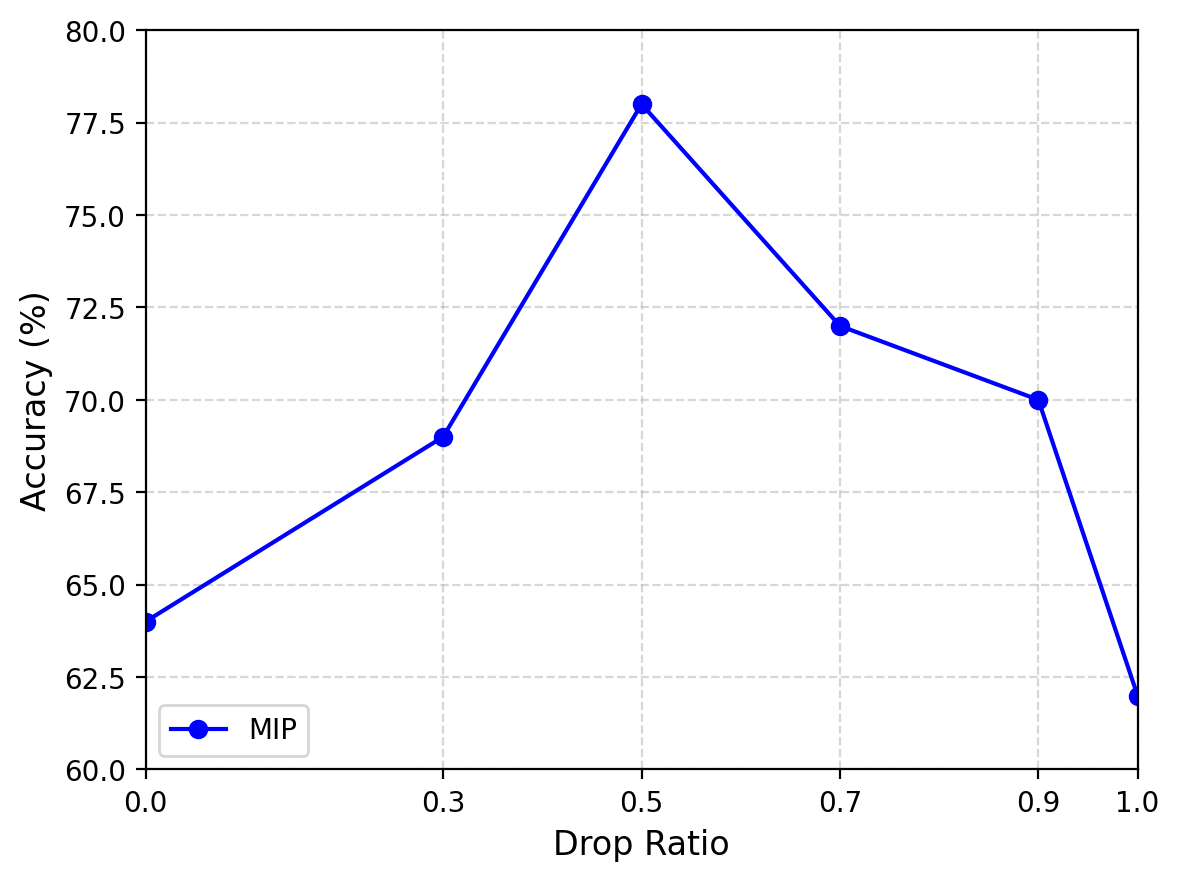}
    \caption{Different drop ratio of the MIP branch.}
\label{fig:drop ratio}
  \end{minipage}
\end{figure}

To leverage the global information of a group captured by the global image and the local-based semantic prior of the MIP, we propose a dual-pathway feature learning approach for group affect prediction. By integrating self-attention learning into the dual-pathway, our approach enables the discovery of spatial-wise discriminative clues both globally and locally. To validate our assumptions, we compute the CKA similarity~\cite{xu2022evo_vit} to measure the similarity of the intermediate token features in each layer and the final CLS token, follwing~\cite{xu2022evo_vit}. 
As illustrated in Figure~\ref{fig:CKA}, both the MIP and global tokens exhibit a positive correlation with the CLS token as the model depth increases. This suggests that both features are valuable for the final classification task and highlights the rationality and necessity of our dual-pathway input design.

Firstly, both the global and MIP image are tokenized into patches, then the class token and a learnable position embedding are added to both branches before the Multi-Scale Transformer Encoder.
In order to balance computational costs, our Transformer Encoders in the two pathways include different numbers (\textit{i.e.}, \textit{G}=6 and \textit{M}=1) inspired by CrossViT~\cite{crossvit}.
The global branch is the large (primary) branch with a coarse patch size (\textit{i.e.}, 12), with a larger embedding size, and more transformer encoders.  
The MIP image is the input of a small (complementary) branch with fine-grained patch size (\textit{i.e.,} 16), fewer encoders, and a smaller embedding size.

As the core of the vision transformer, the Multi-Head Self-Attention (MHSA) 
maps a query vector into a set of key and value vectors~\cite{vit}. 
In the $l$-th Transformer Encoder block in both pathways (a basic Transformer Encoder is shown in Figure~\ref{fig:overall architecture}\textcolor{red}{b}), the output feature map is ${X_l} \in \mathbb{R}^{(N+1)\times d}$.
The attention matrix ${A_l} \in \mathbb{R}^{S\times(N+1)\times (N+1)}$ of MHSA module in the block is computed as:
\begin{equation}
A_{l} = Softmax(\frac{Q_l*K_l^T}{\sqrt{d}}),
\label{CPA_eq}
\end{equation}
where ${Q_l} \in \mathbb{R}^{(N+1)\times d}$ and ${K_l} \in \mathbb{R}^{(N+1)\times d}$  denote the queries and keys projected by ${X_l}$ of self-attention operation in $(l-1)$-th transformer block,
respectively. $S$ represents the number of heads, $T$ is a transpose operator, $d$ is the dimension of the embedding. $N$ is the number of tokenized patches. The input global and MIP images are divided into different patches numbers with different $N$.
The calculation includes $1$ class token in addition to the $N$ tokens.
\subsection{The interaction of global and MIP pathways}
Due to the large amount of information in group-level images, extracting features by either global or MIP branch alone is not sufficient, and cross-attention of discriminative tokens from both can provide complementary effects.

Having obtained global and MIP tokens by the Transformer Encoder, we should consider the dual-path interactions.
To achieve this, we propose the Cross-Patch Attention (CPA) mechanism to establish spatial interactions between the global and MIP images. Firstly, we sort the tokens based on their importance in each path using the Token Ranking Module and create a new query matrix by selecting the top $\eta$ tokens. As depicted in Figure~\ref{fig:drop ratio}, the model's performance, taking the MIP branch as an example~\footnote{The global branch exhibits similar behavior.}, is related to the drop ratio, where dropping a certain percentage of tokens can improve performance. 
Secondly, we compute the Cross-Patch Attention (CPA) using the new query matrix and key-value from the other branch. It is noteworthy that the gradients propagate through both pathways, which allows it to compensate for information loss on global feature compression through global-to-local interaction. As validated in~\cite{zhu2022dual}, such cross-attention can also be viewed as a novel regularization method to regularize attention learning, especially for our small datasets.

\subsubsection{Token Ranking Module}
\label{sec:redundant removal}

For Cross-Patch Attention, we need to remove unimportant contents from the complex group images and keep only the tokens that are relevant.

Our first step is to rank the tokens according to their importance.
While the class token attention map ${A_{cls}} \in \mathbb{R}^{1\times (N+1)}$ reflects the importance of features~\cite{xu2022evo_vit, evit},
we denote the token importance by the similarity scores between the global class token and each patch token:
\begin{equation}
A_{cls} = Softmax(\frac{q_{cls}*K^T}{\sqrt{d}}),
\label{eq:Acls}
\end{equation}
where $q_{cls} \in \mathbb{R}^{1\times d}$ is the query vector of the class token, $K \in \mathbb{R}^{(N+1)\times d}$ is the key vector, and $d$ is the dimension of the embedding.

According to the equation~\ref{eq:Acls}, the weights of each patch correlated with the class token are computed.
In such a way, the $A_{cls}$ reveals how much each patch contributes to the final classification since it captures the global interactions of the class token to all patches.


\subsubsection{Cross-Patch Attention (CPA)}
Based on the importance score of each token, we then construct a newly selected query matrix 
$Q_{sel} \in \mathbb{R}^{(\alpha+1)\times d}$, representing the attentive local embedding, by selecting the top $\alpha$ query vectors that correspond to the top $\alpha$ highest responses in the class attention map. 

In each pathway, the Cross-Patch Attention (CPA) vectors is computed as below,
\begin{equation}
f_{CPA}(Q,K,V) = Softmax(\frac{Q_{sel}*K^T}{\sqrt{d}})V,
\label{CPA_eq}
\end{equation} 
where ${K} \in \mathbb{R}^{(N+1)\times d}$ is the key vector, ${V} \in \mathbb{R}^{(N+1)\times d}$ is the value vector and $d$ is the dimension of the embedding. 

As shown in Figure~\ref{fig:overall architecture}\textcolor{red}{c}, the CPA operation is bidirectional, with each of the two branches providing the selected tokens as the query vector.
Both branches are fused together $C$ times, which means the selected queries are updated in each CPA.


Finally, after repeating $L$ Multi-Scale Transformer Encoder, the class tokens from two branches are combined and sent into a linear layer for prediction. 
The cross-entropy loss is used here for final classification~\cite{crossvit, vit}.



\section{Experiments}
\label{sec:experiments}

\subsection{Implemental details}
Our model is partly pre-trained with the weight of CrossViT~\cite{crossvit} on ImageNet.
We train all our models for 300 epochs (30 warm-up
epochs) on 2 GPUs (RTX 3090 Ti) with a batch size of 64. 
The numbers of Transformer Encoder in the dual path, CPA, and Multi-Scale Transformer Encoder are set as $M$=1, $G$=6, $C$=3, and $L$=1. For the token selection ratios $\alpha$ of (global, MIP), we use (0.5, 0.5) for GAF 3.0 and (1, 0.5) for GroupEmoW throughout the paper. More details can be found in our supplementary materials.

\subsection{Datasets}
\newcommand{\tabincell}[2]{\begin{tabular}{@{}#1@{}}#2\end{tabular}}  

\begin{table*}[tb]
\begin{center}
    \resizebox{1\textwidth}{!}{
    \large
    \begin{tabular}{*{10}{c}}
        \toprule
       Dataset & Label &  Train/Val/Test & Scenario &  Evaluation metric\\
        \midrule
        GroupEmoW~\cite{cohesion} & Neutral, Positive, Negative & 11,127/3,178/1,589 & Specific social events & Acc \\
        GAF 3.0~\cite{cohesion, cohesion_tac} &  Neutral, Positive, Negative & 9,815/4,349/3011 & Specific social events & Acc  \\
        HECO~\cite{ECCV2022_HECO} & \tabincell{c}{Surprise, Excitement, Happiness, Peace,\\Disgust, Anger, Fear, and Sadness}   & 6,570/1,877/938 & \tabincell{c}{In-the-wild images\\that do not limit scenes}   & mAP \\
        GAF Cohesion~\cite{cohesion, cohesion_tac} & 0-3 & 9,815/4,349/3011 & Specific social events & MSE \\
        \bottomrule
        \\
    \end{tabular}
    }    
     \vspace{-0.2cm} 
    \caption{Datasets details used in this paper (Acc = Accuracy, mAP = mean average precision, MSE = mean square error).}
     \label{tab:datasets}
    \end{center}
\end{table*}

In this paper, we utilize several datasets for our experiments, including GroupEmoW~\cite{GroupEmoW_wacv20}, GAF 3.0~\cite{cohesion}, GAF Cohesion~\cite{cohesion, cohesion_tac}, and HECO~\cite{ECCV2022_HECO}, as presented in Table~\ref{tab:datasets}. The GroupEmoW dataset contains images obtained from the web via search engines, using keywords related to various social events such as funerals, birthdays, protests, conferences, and meetings. 
GAF 3.0~\cite{cohesion} and GAF Cohesion~\cite{cohesion, cohesion_tac} datasets were created through web crawling using a diverse set of keywords related to social events, including world cup winners, weddings, family gatherings, laughing clubs, birthday celebrations, siblings, riots, protests, and violence. The cohesion scores in GAF Cohesion range from \textit{strongly agree, agree, disagree} to \textit{strongly disagree}, using a 0-3 scale.

In our experiments, we utilized the MIP ground truth provided by HECO~\cite{ECCV2022_HECO}. However, the ground truth for GAF 3.0 and GroupEmoW was not available. However, since the ground truth for GAF 3.0 and GroupEmoW was not available, we employed the pre-trained POINT model~\cite{POINT} to generate the MIP. The POINT model was trained on the Multi-scene Important People Image Dataset (MS Dataset).~\footnote{To evaluate the effectiveness of the pre-trained POINT model, we labeled the MIP on the GAF 3.0 and GroupEmoW datasets, and further details can be found in the supplementary material.}.



\subsection{Comparison with state-of-the-art group affect methods and ViTs}
\label{sec:comparison}



\begin{figure}
  \centering
     \includegraphics[scale=0.23]{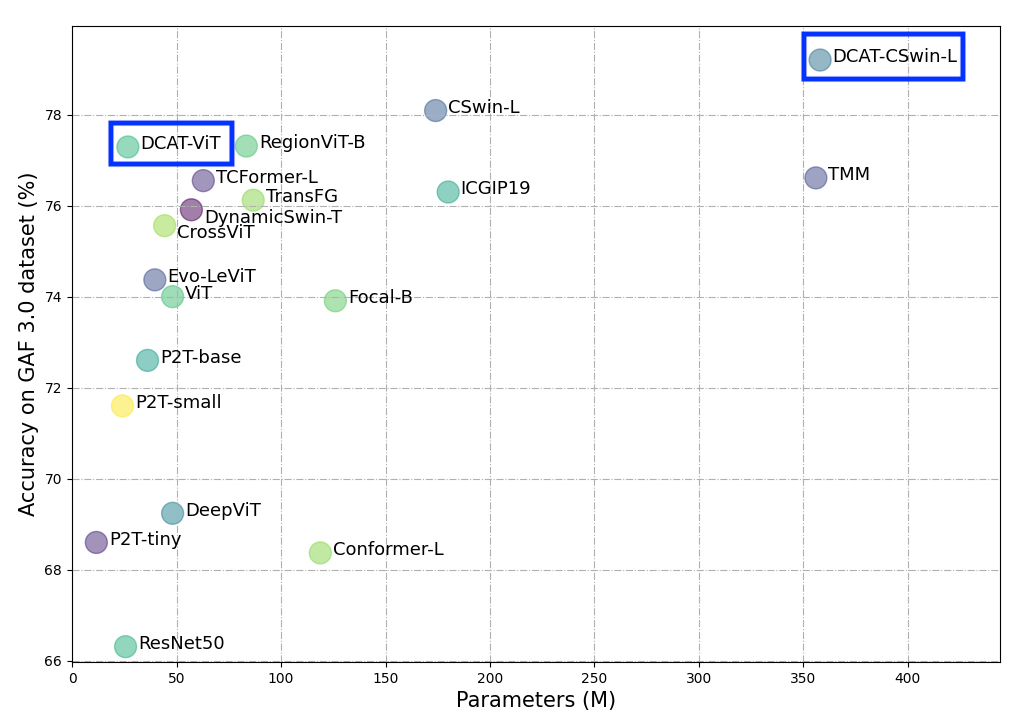}
  \caption{Model parameter and accuracy trade-off comparison on the GAF 3.0 dataset. Our proposed models are highlighted by the blue rectangular box.}
  \label{fig:gaf-param}
\end{figure}


\begin{table}
\centering
\begin{tabular}{lll}
\toprule
Methods         & Acc (\%)  & Source\\ 
\midrule
\textit{Group affect methods} &&\\
ResNet34~\cite{resnet}        &  68.13   &Global, Faces \\
SE-ResNet-50~\cite{hu2018squeeze}     &  69.79  &Global     \\
Efficientnet-b2~\cite{tan2019efficientnet} & 72.33  &Global, Faces   \\
CAER-Net~\cite{lee2019context}       & 80.61   &Global, Faces  \\
GNN~\cite{GroupEmoW_wacv20}         & 82.38&Global     \\
GNN~\cite{GroupEmoW_wacv20}             & 84.62  & Global, Faces, Objects    \\
ConGNN~\cite{wang2022congnn}     & 85.59  &Global, Faces, Objects   \\
WACV 21 ~\cite{groupemow_sota_wacv21}       & 89.36  &Global, Faces   \\
GNN~\cite{GroupEmoW_wacv20}        & 89.93  &Global, Faces   \\
WACV 21~\cite{groupemow_sota_wacv21}          & 90.18   &Global, Faces, Objects  \\
\midrule
\textit{Vision transformers} &&\\
ViT~\cite{vit} & 87.50 &  Global\\
P2T-base~\cite{wu2022p2t} & 86.50 &  Global\\
DeepViT~\cite{zhou2021deepvit} & 83.90 &  Global\\
Conformer-L~\cite{conformer} & 83.35 &  Global\\
Focal-B~\cite{yang2021focal} & 87.98 &  Global\\
CrossViT~\cite{crossvit} & 88.48 &  Global\\
TransFG~\cite{he2022transfg} & 89.47 &  Global\\
RegionViT-B~\cite{chen2021regionvit} & 89.49 &  Global\\
CSwin-L~\cite{dong2022cswin} & 89.90 &  Global\\
DVT~\cite{wang2021DVT}& 87.90 &  Global\\
Evo-LeViT~\cite{xu2022evo_vit}& 87.83 &  Global\\
TCFormer-L~\cite{TCformer}& 89.24 &  Global\\
\midrule
DCAT-ViT & 89.55 & Global, MIP\\

\textbf{DCAT-CSwin-L} & \textbf{90.47} & Global, MIP\\
\bottomrule
\\
\end{tabular}
 \vspace{-0.2cm} 
\caption{Overall Accuracy Comparison on the GroupEmoW dataset. "Source" refers to the types of features used in each paper.}
\label{tab:groupemow}
\end{table}

In this work, we have built our proposed modules based on CrossViT~\cite{crossvit} with vanilla Transformer Encoders~\cite{vit}, which denotes by \textbf{DCAT-ViT}. 
In addition, since there is a series of ViTs without the class token,~\textit{e.g.,} Swin~\cite{liu2021swin} and CSwin~\cite{dong2022cswin}. 
In order to prove the generality of our approach, we have 
added the proposed dual-pathway and CPA module to the CSwin transformer~\cite{dong2022cswin}, which is denoted by \textbf{DCAT-CSwin-L}.

A comparison of the proposed model with the state-of-the-art group affect approaches on GAF 3.0 and GroupEmoW datasets can be found in Figure~\ref{fig:gaf-param} and Table~\ref{tab:groupemow}, respectively. 
Since our proposed DCAT model is based on a vision transformer, we also compare it with recent vision transformer models, such as multi-scale-based (CrossViT~\cite{crossvit}, Focal-B~\cite{yang2021focal}, RegionViT-B~\cite{chen2021regionvit}, TCFormer-L~\cite{TCformer}), CNN-transformer hybrid-based (Conformer-L~\cite{conformer} ), and lightweight-based models (Evo-LeViT~\cite{xu2022evo_vit}, DynamicSwin-T~\cite{rao2021dynamicvit}, DVT~\cite{wang2021DVT}).


\textbf{Comparing with state-of-the-art group affect methods}.
Our DCAT-ViT and DCAT-CSwin-L achieve the best results compared to other group affect recognition methods on the GAF 3.0 dataset. Specifically, our DCAT-ViT achieves 77.29\% accuracy with only 26.70M parameters, which is only about 7.6\% of TMM~\cite{tmm}, while our DCAT-CSwin-L with a larger number of parameters (358M) achieves 79.20\% accuracy. On the GroupEmoW dataset, our DCAT-CSwin-L also achieves the best performance compared to other methods. It is worth noting that our proposed method only needs the global image and corresponding MIP image, while current group affect methods need multiple pre-trained detectors, which may be influenced by imperfect faces and objects. Furthermore, our method learns discriminative clues end-to-end in coordination with self-attention learning.

\textbf{Comparing with single focal attention methods.}
A comparison is made between our dual-pathway attention model and methods specifically designed for local attention (Focal Transformers~\cite{yang2021focal} and TransFG~\cite{he2022transfg}). 
This finding highlights the fact that saliency does not always equate to importance, and that our method's attention mechanism can effectively identify and leverage the most critical affective regions.

Overall, our proposed DCAT-ViT and DCAT-CSwin-L perform better than other group affect recognition models and ViTs in terms of both accuracy and model complexity. 
\subsection{The scalability of our method}
Our proposed architecture can be adapted to accommodate the concept of a \textit{primary agent}, similar to the idea of MIP, which was introduced by HECO~\cite{ECCV2022_HECO}. Specifically, we can modify the input to the MIP branch from the MIP image to the primary agent while keeping the global branch unchanged.

Due to the wide variety of human-object interactions (HOIs) in natural scenes and the fact that interactions of agents’ actions with objects can induce emotional arousal of the primary agent in the scene, a useful approach for emotion recognition in images is to detect HOIs. In previous approaches~\cite{ECCV2022_HECO, lee2019context}, multiple contexts, such as regions of interest (ROIs) and HOIs of all objects, were considered. Our proposed method outperforms previous approaches, as shown in Table~\ref{tab:HECO}. Our proposed method achieved superior performance on the dataset, which may be attributed to the wide variety of HOI categories in various challenging scenarios. Previous methods relied on pre-trained HOI models that were limited by predefined interaction categories. In contrast, our Cross-Patch Attention (CPA) module leverages the ability to learn HOI interactions directly from the data, allowing for greater flexibility and adaptability in recognizing new and diverse HOI categories. 
Moreover, our dual-pathway architecture significantly simplifies the computational process, eliminating the need for a heuristic selection of affective regions.

\begin{table}
\centering
\begin{tabular}{ll}
\hline
Method            & mAP (\%)          \\
\hline
GCNs~\cite{GCN}              & 48.84           \\
Depth~\cite{li2018megadepth}             & 49.02           \\
OpenFace~\cite{baltruvsaitis2016openface}+OpenPose~\cite{openpose} & 50.28           \\
R-FCN~\cite{RFCN}+HOI-Net~\cite{liao2020ppdm_HOI}     & 50.34           \\
HECO~\cite{ECCV2022_HECO}      & 50.24           \\
\hline
\textbf{DCAT-ViT}     & \textbf{53.49}  \\
\textbf{DCAT-CSwin-L}     & \textbf{58.06}  \\
\hline 
\\
\end{tabular}
 \vspace{-0.2cm} 
\caption{Comparison on HECO dataset.}
\label{tab:HECO}
\end{table}

 




\subsection{Transfer learning for group cohesion estimation}
\begin{figure}
  \centering
     \includegraphics[scale=0.85]{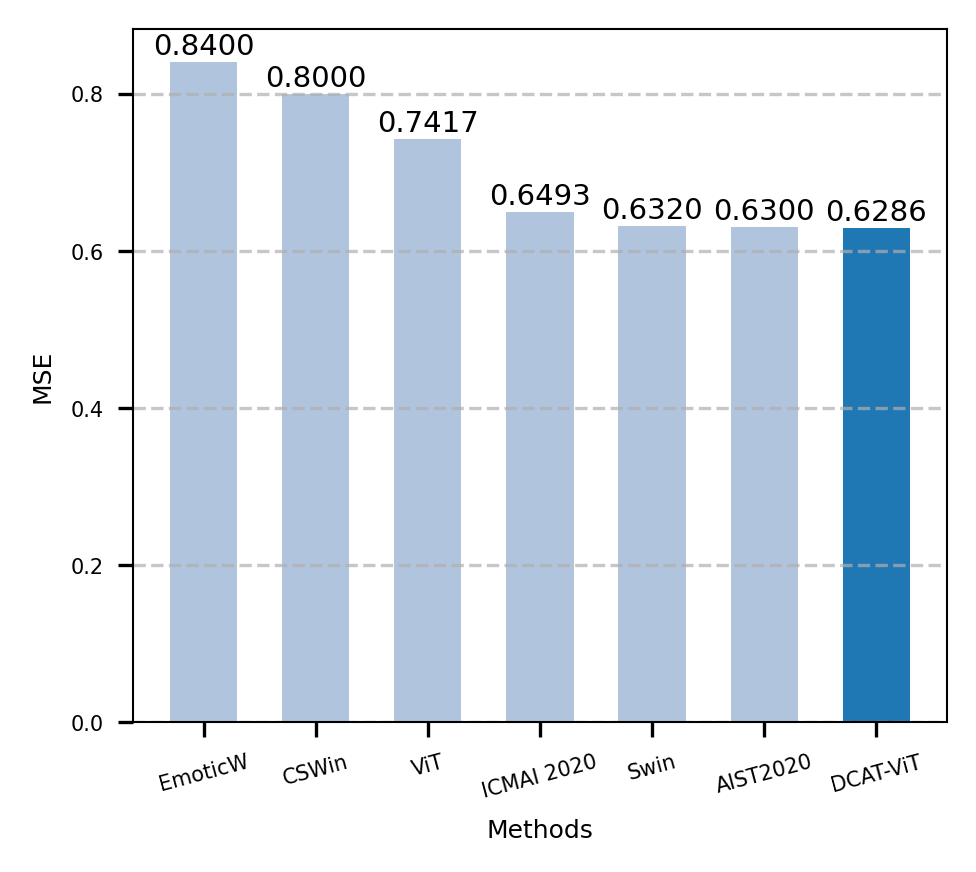}
  \caption{Group cohesion score comparison in the GAF Cohesion dataset (MSE is the evaluation metric, the lower the better).}
  \label{fig:cohesion}
\end{figure}

Group cohesion is an affective phenomenon and a measure of agreement between members. 
Since cohesion and emotion (or valence) are intertwined in social science and affective computing, our DCAT for group valence prediction can be extented to group cohesion estimation.
In this work, we finetune our model pre-trained on GroupEmoW for the GAF Cohesion dataset.
The results in Figure~\ref{fig:cohesion} assure that our models still have good generalization ability rather than only fit to group affect datasets compared to other group affect recognition methods, \textit{e.g.}, EmoticW~\cite{cohesion}, ICMAI2020~\cite{zou2020joint}, AIST2020~\cite{gavrikov2020efficient}, as well as the ViTs, i.e., ViT~\cite{vit}, CSwin~\cite{dong2022cswin}, Swin transformer~\cite{liu2021swin}.

\subsection{Ablation study}
\label{sec:ablation study}

\textbf{The effectiveness of each module.}
The effectiveness of each module proposed in our work, \textit{i.e.}, dual-pathway input, CPA module, Token Ranking Module, is shown in Table~\ref{tab:ablation}.
For both the DCAT-ViT and DCAT-CSwin-L, each of the modules contributes to our overall results.


\begin{table}
\begin{center}
\begin{tabular}{lll}
\toprule
Module                                       & GAF 3.0 & GroupEmoW  \\
\midrule
CrossVit                                     & 75.56            & 88.48               \\
CrossVit+A                                 & 75.79            & 88.86               \\
CrossVit+A+B                 & 76.21            & 88.96               \\
CrossVit+A+B+C & \textbf{77.29 }           & \textbf{89.55}\\
\bottomrule
CSwin-L                                     & 78.09            & 89.90              \\
CSwin-L+A & 78.93        & 90.12\\
CSwin-L+A+B & \textbf{79.20}           & \textbf{90.47}\\
\bottomrule
\\
\end{tabular}
\caption{The accuracy (\%) of each proposed module. \textit{A}, \textit{B}, and \textit{C} represent dual-pathway input, CPA module, and Token Ranking Module, respectively. The downsampling operation in each stage of the CSWin-L can be regarded as a token selection mechanism, so the Token Ranking Module is not included in it. }
\label{tab:ablation}
\end{center}
\end{table}

\textbf{The effectiveness of dual-input.}
Hence, we should also dynamically consider the global affective context while learning group affect, as validated in Table~\ref{tab:dual-input}, where MIP and global inputs together perform better than either alone.

\begin{table}
\begin{center}
\begin{tabular}{lll}
\toprule
  Input   & GAF 3.0&GroupEmoW   \\
\midrule
Only MIP     &   68.02  & 80.08          \\
Only global   &    75.56  & 88.48           \\
\textbf{MIP + global} &\textbf{77.29} & \textbf{89.55}            \\
\bottomrule
\\
\end{tabular}
\caption{Effectiveness of dual-input on GAF 3.0~\cite{gaf3.0} and GroupEmoW~\cite{groupemow_sota_wacv21} dataset (evaluation metric is ACC (\%)).}
\label{tab:dual-input}
\end{center}
\end{table}

\textbf{Cross-Cls-Attention (CCA) vs Our proposed Cross-Patch-Attention (CPA).}
It is worth comparing between Cross-Cls-Attention (CCA) proposed in CrossViT~\cite{crossvit} and our proposed CPA (as provided in Table~\ref{tab:cca_cpa}). 
The CCA only exchanges the class token in the cross-attention mechanism, which limits its ability to capture global interactions between class tokens and patches in each branch. In complex scenarios, such as those encountered in group affect recognition, our proposed CPA approach is more effective for modeling the interdependencies between global and MIP features.


\begin{table}
\centering
\begin{tabular}{lll}
\toprule
     & GAF 3.0&GroupEmoW   \\
\midrule
CCA     &   75.41\%  & 87.86\%           \\
\textbf{CPA}   &    \textbf{77.29}\%  & \textbf{89.55}\%           \\
\bottomrule
\\
\end{tabular}
 \vspace{-0.2cm} 
\caption{Comparison between the CCA and our proposed CPA (using the DCAT-ViT) on GAF 3.0 and GroupEmoW dataset.}
\label{tab:cca_cpa}
\end{table}

\subsection{Visualization of the interaction of global and MIP pathways}

\begin{figure}
  \centering
    \includegraphics[scale=0.5]{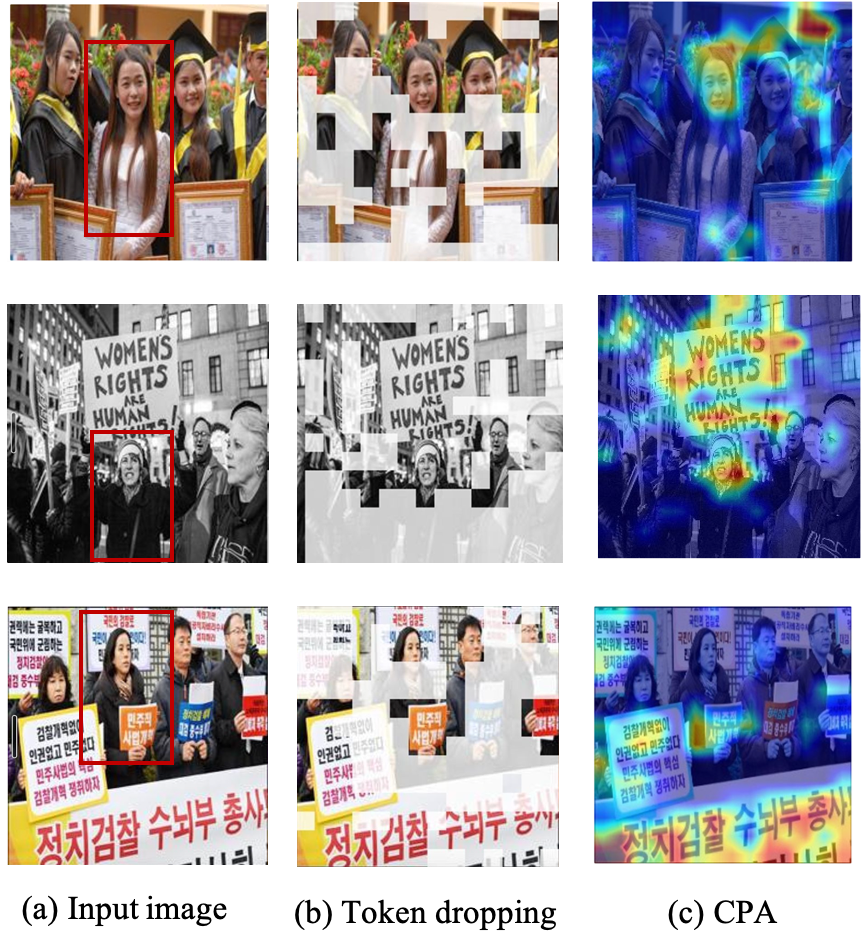}
    \vspace{-0.2cm}  
  \caption{The visualization of the interaction of global and MIP.}
  \label{fig:visualization of CPA}
\end{figure}

The interaction of the global and MIP pathways, including token dropping and CPA, is demonstrated in Figure~\ref{fig:visualization of CPA}. As depicted in column (b), irrelevant background and other facial details are eliminated, while the MIP region and significant background regions (such as banners in protest scenes) are preserved for cross-patch attention. This demonstrates the effectiveness of our MIP-global interaction and highlights its efficacy in capturing discriminative information.


\subsection{Case study: the relationship between group size and performance}

\begin{figure}
  \centering
     \includegraphics[scale=0.55]{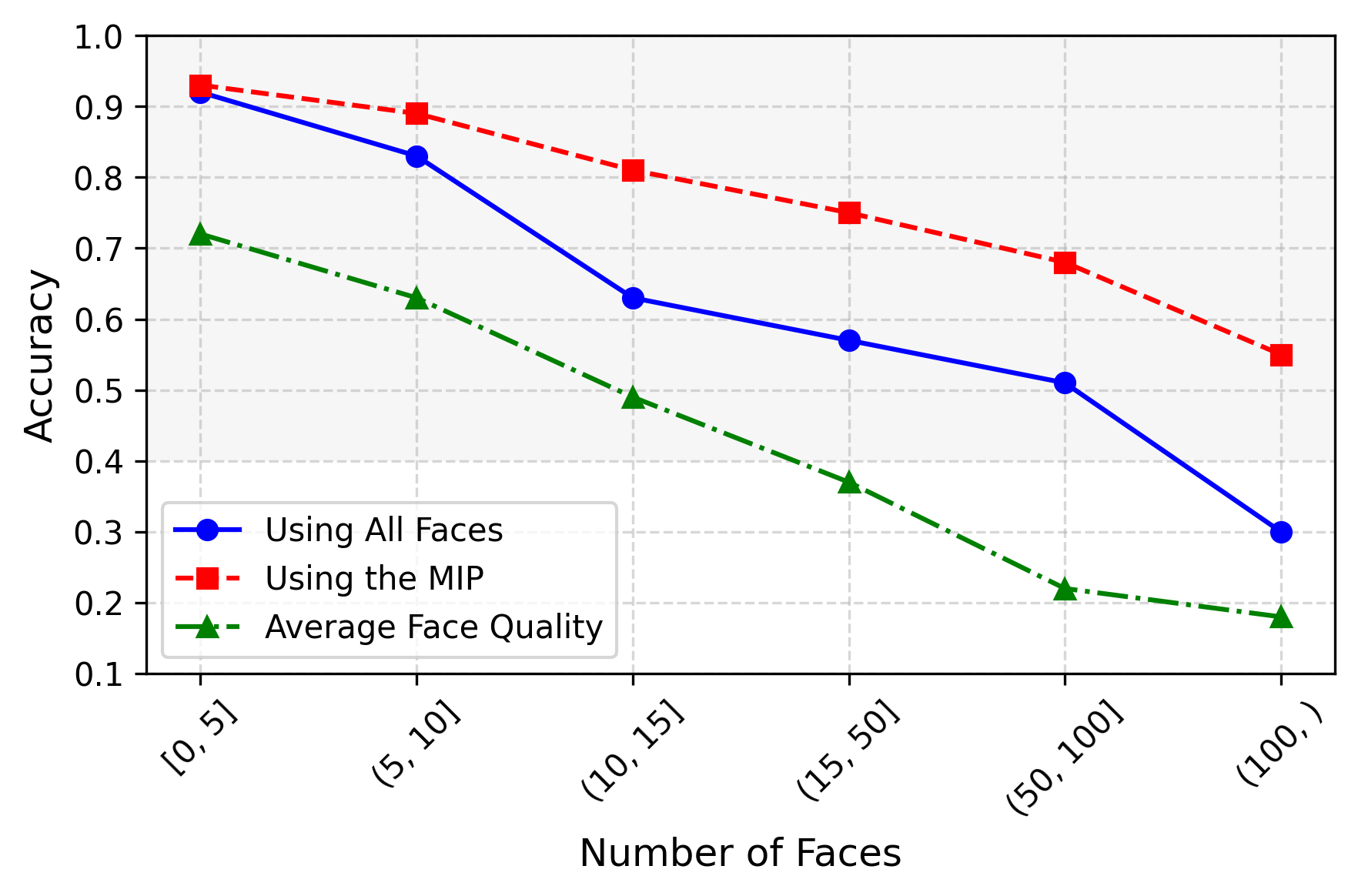}
      \vspace{-0.2cm} 
  \caption{The performance of average face quality, using all faces and using the MIP under different face count intervals in the image. The GroupEmoW dataset is used here.}
  \label{fig:quality_face_number}
\end{figure}

Previous studies have suggested that distinguishing between the \textit{negative} and \textit{neutral} categories in a group can be challenging~\cite{tmm}. In this paper, we present new insights into this issue. We first analyze the number of individuals in each image and find that as the number of individuals increases, the average quality of their facial features decreases~\footnote{We quantify face quality based on~\cite{quality}.}, as illustrated in Figure~\ref{fig:quality_face_number}. 
This indicates that facial details become increasingly blurred and difficult to extract, posing a significant challenge, particularly when trying to distinguish between negative and neutral categories, which are inherently difficult to distinguish.
In contrast to previous works that typically average features across all detected faces~\cite{tmm}, our approach selectively focuses on the most informative facial region, i.e., the MIP, and outperforms models that utilize all faces in images with larger populations. This approach avoids interference from low-quality faces and effectively interacts between the features of the MIP and those of the entire global image through our proposed CPA module, enhancing the accuracy of our results.
Therefore, our approach yields higher accuracy for distinguishing between the negative and neutral categories compared to previous methods. The exact numerical results can be found in the supplementary material.

\section{Conclusions}
In this work, by incorporating the MIP concept, we propose a dual-pathway vision transformer model, the Dual-branch Cross-Patch Attention Transformer (DCAT) for group affect recognition.
Moreover, we propose the CPA module so that the important tokens from both paths can be interacted with.
In terms of accuracy and parameters, the proposed DCAT outperforms both state-of-the-art group affect models and vision transformer models. It is also the first study to explore group affect recognition beyond CNNs. Moreover, our proposed model can be utilized in other group-level affect tasks, \textit{i.e.}, group cohesion analysis.



 \section*{Acknowledgement}
\label{sec:ack} 
This work was supported in part by the National Science and Technology Council of Taiwan under Grants NSTC-109-2223-E-002-005-MY3, NSTC-112-2628-E-002-033-MY4 and NSTC-112-2634-F-002-002-MBK.

\clearpage
\clearpage
{\small
\bibliographystyle{ieee_fullname}


\begin{thebibliography}{10}\itemsep=-1pt

\bibitem{baltruvsaitis2016openface}
Tadas Baltru{\v{s}}aitis, Peter Robinson, and Louis-Philippe Morency.
\newblock Openface: an open source facial behavior analysis toolkit.
\newblock In {\em 2016 IEEE winter conference on applications of computer vision (WACV)}, pages 1--10. IEEE, 2016.

\bibitem{groupaffect}
Sigal~G Barsade and Donald~E Gibson.
\newblock Group affect: Its influence on individual and group outcomes.
\newblock {\em Current Directions in Psychological Science}, 21(2):119--123, 2012.

\bibitem{openpose}
Zhe Cao, Tomas Simon, Shih-En Wei, and Yaser Sheikh.
\newblock Realtime multi-person 2d pose estimation using part affinity fields.
\newblock In {\em Proceedings of the IEEE conference on computer vision and pattern recognition}, pages 7291--7299, 2017.

\bibitem{chen2021regionvit}
Chun-Fu Chen, Rameswar Panda, and Quanfu Fan.
\newblock Regionvit: Regional-to-local attention for vision transformers.
\newblock {\em arXiv preprint arXiv:2106.02689}, 2021.

\bibitem{crossvit}
Chun-Fu~Richard Chen, Quanfu Fan, and Rameswar Panda.
\newblock Crossvit: Cross-attention multi-scale vision transformer for image classification.
\newblock In {\em Proceedings of the IEEE/CVF international conference on computer vision}, pages 357--366, 2021.

\bibitem{RFCN}
Jifeng Dai, Yi Li, Kaiming He, and Jian Sun.
\newblock R-fcn: Object detection via region-based fully convolutional networks.
\newblock {\em Advances in neural information processing systems}, 29, 2016.

\bibitem{dong2022cswin}
Xiaoyi Dong, Jianmin Bao, Dongdong Chen, Weiming Zhang, Nenghai Yu, Lu Yuan, Dong Chen, and Baining Guo.
\newblock Cswin transformer: A general vision transformer backbone with cross-shaped windows.
\newblock In {\em Proceedings of the IEEE/CVF Conference on Computer Vision and Pattern Recognition}, pages 12124--12134, 2022.

\bibitem{vit}
Alexey Dosovitskiy, Lucas Beyer, Alexander Kolesnikov, Dirk Weissenborn, Xiaohua Zhai, Thomas Unterthiner, Mostafa Dehghani, Matthias Minderer, Georg Heigold, Sylvain Gelly, Jakob Uszkoreit, and Neil Houlsby.
\newblock An image is worth 16x16 words: Transformers for image recognition at scale.
\newblock {\em ICLR}, 2021.

\bibitem{tmm}
Katsuya Fujii, Daisuke Sugimura, and Takayuki Hamamoto.
\newblock Hierarchical group-level emotion recognition.
\newblock {\em IEEE Transactions on Multimedia}, 23:3892--3906, 2020.

\bibitem{gavrikov2020efficient}
Ilya Gavrikov and Andrey~V Savchenko.
\newblock Efficient group-based cohesion prediction in images using facial descriptors.
\newblock In {\em International Conference on Analysis of Images, Social Networks and Texts}, pages 140--148. Springer, 2020.

\bibitem{groupleader_1}
Jennifer~M George.
\newblock Leader positive mood and group performance: The case of customer service.
\newblock {\em Journal of Applied Social Psychology}, 25(9):778--794, 1995.

\bibitem{saliency_importa}
Shreya Ghosh and Abhinav Dhall.
\newblock Role of group level affect to find the most influential person in images.
\newblock In {\em Proceedings of the European Conference on Computer Vision (ECCV) Workshops}, pages 0--0, 2018.

\bibitem{cohesion}
Shreya Ghosh, Abhinav Dhall, Nicu Sebe, and Tom Gedeon.
\newblock Predicting group cohesiveness in images.
\newblock In {\em 2019 International Joint Conference on Neural Networks (IJCNN)}, pages 1--8. IEEE, 2019.

\bibitem{gaf3.0}
S. {Ghosh}, A. {Dhall}, N. {Sebe}, and T. {Gedeon}.
\newblock Predicting group cohesiveness in images.
\newblock In {\em 2019 International Joint Conference on Neural Networks (IJCNN)}, pages 1--8, 2019.

\bibitem{cohesion_tac}
Shreya Ghosh, Abhinav Dhall, Nicu Sebe, and Tom Gedeon.
\newblock Automatic prediction of group cohesiveness in images.
\newblock {\em IEEE Transactions on Affective Computing}, 2020.

\bibitem{GroupEmoW_wacv20}
Xin Guo, Luisa Polania, Bin Zhu, Charles Boncelet, and Kenneth Barner.
\newblock Graph neural networks for image understanding based on multiple cues: Group emotion recognition and event recognition as use cases.
\newblock In {\em Proceedings of the IEEE/CVF Winter Conference on Applications of Computer Vision}, pages 2921--2930, 2020.

\bibitem{he2022transfg}
Ju He, Jie-Neng Chen, Shuai Liu, Adam Kortylewski, Cheng Yang, Yutong Bai, and Changhu Wang.
\newblock Transfg: A transformer architecture for fine-grained recognition.
\newblock In {\em Proceedings of the AAAI Conference on Artificial Intelligence}, volume~36, pages 852--860, 2022.

\bibitem{resnet}
Kaiming He, Xiangyu Zhang, Shaoqing Ren, and Jian Sun.
\newblock Deep residual learning for image recognition.
\newblock In {\em Proceedings of the IEEE conference on computer vision and pattern recognition}, pages 770--778, 2016.

\bibitem{MIP_hong2020learning}
Fa-Ting Hong, Wei-Hong Li, and Wei-Shi Zheng.
\newblock Learning to detect important people in unlabelled images for semi-supervised important people detection.
\newblock In {\em Proceedings of the IEEE/CVF Conference on Computer Vision and Pattern Recognition}, pages 4146--4154, 2020.

\bibitem{hu2018squeeze}
Jie Hu, Li Shen, and Gang Sun.
\newblock Squeeze-and-excitation networks.
\newblock In {\em Proceedings of the IEEE conference on computer vision and pattern recognition}, pages 7132--7141, 2018.

\bibitem{kelly2001mood}
Janice~R Kelly and Sigal~G Barsade.
\newblock Mood and emotions in small groups and work teams.
\newblock {\em Organizational behavior and human decision processes}, 86(1):99--130, 2001.

\bibitem{groupemow_sota_wacv21}
Ahmed~Shehab Khan, Zhiyuan Li, Jie Cai, and Yan Tong.
\newblock Regional attention networks with context-aware fusion for group emotion recognition.
\newblock In {\em Proceedings of the IEEE/CVF Winter Conference on Applications of Computer Vision}, pages 1150--1159, 2021.

\bibitem{GCN}
Thomas~N Kipf and Max Welling.
\newblock Semi-supervised classification with graph convolutional networks.
\newblock {\em arXiv preprint arXiv:1609.02907}, 2016.

\bibitem{knapp2011physiological}
R~Benjamin Knapp, Jonghwa Kim, and Elisabeth Andr{\'e}.
\newblock Physiological signals and their use in augmenting emotion recognition for human--machine interaction.
\newblock In {\em Emotion-oriented systems}, pages 133--159. Springer, 2011.

\bibitem{lee2019context}
Jiyoung Lee, Seungryong Kim, Sunok Kim, Jungin Park, and Kwanghoon Sohn.
\newblock Context-aware emotion recognition networks.
\newblock In {\em Proceedings of the IEEE/CVF international conference on computer vision}, pages 10143--10152, 2019.

\bibitem{POINT}
Wei-Hong Li, Fa-Ting Hong, and Wei-Shi Zheng.
\newblock Learning to learn relation for important people detection in still images.
\newblock In {\em Proceedings of the IEEE/CVF Conference on Computer Vision and Pattern Recognition}, pages 5003--5011, 2019.

\bibitem{MIP_li2018personrank}
Wei-Hong Li, Benchao Li, and Wei-Shi Zheng.
\newblock Personrank: Detecting important people in images.
\newblock In {\em 2018 13th IEEE International Conference on Automatic Face \& Gesture Recognition (FG 2018)}, pages 234--241. IEEE, 2018.

\bibitem{li2018megadepth}
Zhengqi Li and Noah Snavely.
\newblock Megadepth: Learning single-view depth prediction from internet photos.
\newblock In {\em Proceedings of the IEEE conference on computer vision and pattern recognition}, pages 2041--2050, 2018.

\bibitem{evit}
Youwei Liang, Chongjian Ge, Zhan Tong, Yibing Song, Jue Wang, and Pengtao Xie.
\newblock Not all patches are what you need: Expediting vision transformers via token reorganizations.
\newblock In {\em The International Conference on Learning Representations (ICLR)}, 2022.

\bibitem{liao2020ppdm_HOI}
Yue Liao, Si Liu, Fei Wang, Yanjie Chen, Chen Qian, and Jiashi Feng.
\newblock Ppdm: Parallel point detection and matching for real-time human-object interaction detection.
\newblock In {\em Proceedings of the IEEE/CVF Conference on Computer Vision and Pattern Recognition}, pages 482--490, 2020.

\bibitem{liu2021swin}
Ze Liu, Yutong Lin, Yue Cao, Han Hu, Yixuan Wei, Zheng Zhang, Stephen Lin, and Baining Guo.
\newblock Swin transformer: Hierarchical vision transformer using shifted windows.
\newblock In {\em Proceedings of the IEEE/CVF International Conference on Computer Vision}, pages 10012--10022, 2021.

\bibitem{conformer}
Zhiliang Peng, Wei Huang, Shanzhi Gu, Lingxi Xie, Yaowei Wang, Jianbin Jiao, and Qixiang~Ye Conformer.
\newblock Local features coupling global representations for visual recognition. in 2021 ieee.
\newblock In {\em CVF International Conference on Computer Vision, ICCV}, pages 357--366, 2021.

\bibitem{polo2016group}
Claire Polo, Kristine Lund, Christian Plantin, and Gerald~P Niccolai.
\newblock Group emotions: The social and cognitive functions of emotions in argumentation.
\newblock {\em International Journal of Computer-Supported Collaborative Learning}, 11(2):123--156, 2016.

\bibitem{qi2021zero}
Fan Qi, Xiaoshan Yang, and Changsheng Xu.
\newblock Zero-shot video emotion recognition via multimodal protagonist-aware transformer network.
\newblock In {\em Proceedings of the 29th ACM International Conference on Multimedia}, pages 1074--1083, 2021.

\bibitem{MIP_ramanathan2016detecting}
Vignesh Ramanathan, Jonathan Huang, Sami Abu-El-Haija, Alexander Gorban, Kevin Murphy, and Li Fei-Fei.
\newblock Detecting events and key actors in multi-person videos.
\newblock In {\em Proceedings of the IEEE conference on computer vision and pattern recognition}, pages 3043--3053, 2016.

\bibitem{rao2021dynamicvit}
Yongming Rao, Wenliang Zhao, Benlin Liu, Jiwen Lu, Jie Zhou, and Cho-Jui Hsieh.
\newblock Dynamicvit: Efficient vision transformers with dynamic token sparsification.
\newblock {\em Advances in neural information processing systems}, 34:13937--13949, 2021.

\bibitem{groupleader_2}
Thomas Sy, St{\'e}phane C{\^o}t{\'e}, and Richard Saavedra.
\newblock The contagious leader: impact of the leader's mood on the mood of group members, group affective tone, and group processes.
\newblock {\em Journal of applied psychology}, 90(2):295, 2005.

\bibitem{tan2019efficientnet}
Mingxing Tan and Quoc Le.
\newblock Efficientnet: Rethinking model scaling for convolutional neural networks.
\newblock In {\em International conference on machine learning}, pages 6105--6114. PMLR, 2019.

\bibitem{quality}
Philipp Terh{\"{o}}rst, Jan~Niklas Kolf, Naser Damer, Florian Kirchbuchner, and Arjan Kuijper.
\newblock {SER-FIQ:} unsupervised estimation of face image quality based on stochastic embedding robustness.
\newblock In {\em 2020 {IEEE/CVF} Conference on Computer Vision and Pattern Recognition, {CVPR} 2020, Seattle, WA, USA, June 13-19, 2020}, pages 5650--5659. {IEEE}, 2020.

\bibitem{wang2021DVT}
Yulin Wang, Rui Huang, Shiji Song, Zeyi Huang, and Gao Huang.
\newblock Not all images are worth 16x16 words: Dynamic transformers for efficient image recognition.
\newblock {\em Advances in Neural Information Processing Systems}, 34:11960--11973, 2021.

\bibitem{wang2022congnn}
Yu Wang, Shunping Zhou, Yuanyuan Liu, Kunpeng Wang, Fang Fang, and Haoyue Qian.
\newblock Congnn: Context-consistent cross-graph neural network for group emotion recognition in the wild.
\newblock {\em Information Sciences}, 610:707--724, 2022.

\bibitem{wu2022p2t}
Yu-Huan Wu, Yun Liu, Xin Zhan, and Ming-Ming Cheng.
\newblock P2t: Pyramid pooling transformer for scene understanding.
\newblock {\em IEEE Transactions on Pattern Analysis and Machine Intelligence}, 2022.

\bibitem{xu2022evo_vit}
Yifan Xu, Zhijie Zhang, Mengdan Zhang, Kekai Sheng, Ke Li, Weiming Dong, Liqing Zhang, Changsheng Xu, and Xing Sun.
\newblock Evo-vit: Slow-fast token evolution for dynamic vision transformer.
\newblock In {\em Proceedings of the AAAI Conference on Artificial Intelligence}, volume~36, pages 2964--2972, 2022.

\bibitem{ECCV2022_HECO}
Dingkang Yang, Shuai Huang, Shunli Wang, Yang Liu, Peng Zhai, Liuzhen Su, Mingcheng Li, and Lihua Zhang.
\newblock Emotion recognition for multiple context awareness.
\newblock {\em ECCV}, 2022.

\bibitem{yang2021focal}
Jianwei Yang, Chunyuan Li, Pengchuan Zhang, Xiyang Dai, Bin Xiao, Lu Yuan, and Jianfeng Gao.
\newblock Focal self-attention for local-global interactions in vision transformers, 2021.

\bibitem{yao2022dual}
Ting Yao, Yehao Li, Yingwei Pan, Yu Wang, Xiao-Ping Zhang, and Tao Mei.
\newblock Dual vision transformer.
\newblock {\em arXiv preprint arXiv:2207.04976}, 2022.

\bibitem{TCformer}
Wang Zeng, Sheng Jin, Wentao Liu, Chen Qian, Ping Luo, Wanli Ouyang, and Xiaogang Wang.
\newblock Not all tokens are equal: Human-centric visual analysis via token clustering transformer.
\newblock In {\em Proceedings of the IEEE/CVF Conference on Computer Vision and Pattern Recognition}, pages 11101--11111, 2022.

\bibitem{zhou2021deepvit}
Daquan Zhou, Bingyi Kang, Xiaojie Jin, Linjie Yang, Xiaochen Lian, Zihang Jiang, Qibin Hou, and Jiashi Feng.
\newblock Deepvit: Towards deeper vision transformer.
\newblock {\em arXiv preprint arXiv:2103.11886}, 2021.

\bibitem{zhu2022dual}
Haowei Zhu, Wenjing Ke, Dong Li, Ji Liu, Lu Tian, and Yi Shan.
\newblock Dual cross-attention learning for fine-grained visual categorization and object re-identification.
\newblock In {\em Proceedings of the IEEE/CVF Conference on Computer Vision and Pattern Recognition}, pages 4692--4702, 2022.

\bibitem{zou2020joint}
Bochao Zou, Zhifeng Lin, Haoyi Wang, Yingxue Wang, Xiangwen Lyu, and Haiyong Xie.
\newblock Joint prediction of group-level emotion and cohesiveness with multi-task loss.
\newblock In {\em Proceedings of the 2020 5th International Conference on Mathematics and Artificial Intelligence}, pages 24--28, 2020.

\end{thebibliography}
}
\end{document}